\documentclass[twoside,11pt]{article}

\usepackage{blindtext}

\usepackage[preprint]{jmlr2e}

\usepackage{lastpage}

\ShortHeadings{TorchKM: A GPU‑Oriented Library for Kernel Learning and Model Selection}{Zhang, Jia, Ding, and Wang}
\firstpageno{1}
\usepackage{booktabs}
\usepackage{graphicx} 
\usepackage{amsmath}
\usepackage{caption}
\usepackage{enumitem}
\usepackage{multirow}
\usepackage{algorithm}
\usepackage{algorithmic}
\usepackage{xcolor}
\usepackage{pifont}

\newcommand{\cmark}{\textcolor{green!60!black}{\ding{51}}}
\newcommand{\xmark}{\textcolor{red!75!black}{\ding{55}}}

\definecolor{codebg}{gray}{0.97}

\usepackage{xcolor}
\usepackage{listings}

\definecolor{codebg}{RGB}{248,248,248}
\definecolor{codeblue}{RGB}{0,0,180}
\definecolor{codegreen}{RGB}{0,128,0}
\definecolor{codered}{RGB}{180,0,0}

\lstdefinestyle{torchkmstyle}{
    language=Python,
    basicstyle=\ttfamily\small,
    backgroundcolor=\color{codebg},
    keywordstyle=\color{codeblue}\bfseries,
    commentstyle=\color{codegreen},
    stringstyle=\color{codered},
    showstringspaces=false,
    columns=fullflexible,
    keepspaces=true,
    breaklines=true,
    frame=none,
    xleftmargin=0pt,
    aboveskip=0.75em,
    belowskip=0.75em,
    morekeywords={True,False,None}
}

\lstnewenvironment{pythoncode}
{\lstset{style=torchkmstyle}}
{}

\DeclareMathOperator*{\argmin}{argmin}%
\newcommand{\bs}{\boldsymbol}%

\begin{document}

\title{TorchKM: A GPU‑Oriented Library for Kernel Learning and Model Selection}

\author{\name Yikai Zhang \email yikai-zhang@uiowa.edu \\
       \addr Department of Statistics and Actuarial Science, University of Iowa, Iowa City, IA 52242, USA
       \AND
       \name Gaoxiang Jia\email jiagaoxiang@gmail.com \\
       \addr Individual Researcher
       \AND
        \name Jie Ding\email jie@morphmind.ai \\
       \addr AIScientists, Inc. (MorphMind), 245 Main Street, Cambridge, MA 02142, USA
       \AND
       \name Boxiang Wang \email boxiang-wang@uiowa.edu \\
       \addr Department of Statistics and Actuarial Science, University of Iowa, Iowa City, IA 52242, USA}


\maketitle

\begin{abstract}
\texttt{TorchKM} is an open-source library for kernel machines, including support vector machines, kernel logistic regression, and kernel quantile regression, with GPU acceleration. The library features a scikit-learn-style API and is designed to exploit GPU-friendly linear algebra, accelerating the full training and model-selection pipeline through intelligent reuse of matrix operations. Benchmarks show competitive predictive performance with substantial speedups over standard baselines. The efficiency and programmable design also make \texttt{TorchKM} a kernel-learning component for AI-driven workflows. Code and documentation are available at \url{https://github.com/YikaiZhang95/torchkm}, and the package can be easily installed via PyPI.
\end{abstract}


\section{Introduction}

Kernel machines, including support vector machines \citep[SVMs,][]{cortes1995support}, have been a fundamental class of methods in machine learning for several decades. They are widely recognized for their strong predictive performance, a convex optimization framework with unique global solutions, and a rigorous theoretical foundation. Although deep neural networks currently dominate fields such as computer vision and natural language processing, kernel machines remain highly competitive in structured-data applications and continue to define the state-of-the-art in many fields, as evidenced by recent comprehensive evaluations in \cite{cervantes2020survey}.

In modern practice, however, kernel machines are often relegated to the ``back seat'' of the machine learning toolbox, and the primary barrier is computational cost. The performance of an SVM, for instance, is sensitive to hyperparameter choices, but identifying a good configuration can be prohibitively expensive. Classical implementations, such as \texttt{LIBSVM} \citep{chang2011libsvm}, \texttt{scikit-learn} \citep{pedregosa2011scikit}, and \texttt{kernlab} \citep{karatzoglou2004kernlab}, follow a standard paradigm in which training and model selection are treated as separate stages. Although fitting a model at each tuning parameter can be made efficient through specialized algorithms, such as sequential minimal optimization \citep{platt1999fast}, tuning is still typically carried out in an outer loop through repeated refitting over a parameter grid, often at substantial computation cost. As such, kernel methods are often under-tuned in practice, and thus may not achieve their full predictive potential.

One promising way to reduce computational cost is to exploit advances in modern computing hardware, particularly graphics processing unit (GPU) acceleration \citep{nickolls2008scalable}. An important breakthrough in this direction was \texttt{ThunderSVM} \citep{wen2018thundersvm, jiang2021parallel}, which substantially accelerated kernel SVM training on GPUs. Despite this important progress, end-to-end kernel-machine workflows can further benefit from acceleration beyond single model fits. The main reason is that repeated matrix operations across cross-validation folds and tuning parameters can often dominate the overall computational cost. Reducing the computational overhead for tuning is therefore essential for fully exploiting GPU acceleration in kernel-machine computation. 

In this paper, we present \texttt{TorchKM}, an open-source, GPU-oriented library for efficient end-to-end kernel learning with integrated model selection. \texttt{TorchKM} supports a broad range of methods for both regression and classification, including quantile regression, logistic regression, SVM, and distance-weighted discrimination (DWD). It provides a user-friendly scikit-learn-style interface. Rather than directly porting an existing CPU solver to the GPU, \texttt{TorchKM} is built through algorithm--hardware co-design: its computational algorithms are designed to exploit GPU-friendly linear algebra. To achieve this goal, at the core of \texttt{TorchKM} are two key algorithmic ideas, an exact cross-validation formula and a spectral algorithm, which together speed up the full training and tuning pipeline and still produce \textit{exact} solutions rather than approximations. 

Beyond conventional interactive use, \texttt{TorchKM} is also well suited as a kernel-learning tool for agentic and automated machine-learning systems. Its integrated train-and-tune pipeline replaces an external search loop with a single call, exact cross-validation yields a reliable model-selection signal, and calibrated classification probabilities (see Section A.7) provide the kind of output such systems can act on directly in downstream decisions.

\begin{table}[t]
\centering
\caption{Comparison of representative kernel-learning libraries. ThunderSVM is known as a GPU-accelerated SVM library; TorchKM focuses on integrated model-selection pipeline.}
\label{tab:library_comparison}
\renewcommand{\arraystretch}{0.94}
\setlength{\tabcolsep}{6pt}
\begin{tabular}{p{5.6cm}ccc}
\toprule
\textbf{Features} & \texttt{scikit-learn} & \texttt{ThunderSVM} & \texttt{TorchKM} \\
\midrule
GPU acceleration & \xmark & \cmark & \cmark \\
Integrated training and tuning & \xmark & \xmark & \cmark \\
Non-SVM losses & \cmark & \xmark & \cmark \\
Probability calibration & \cmark & \cmark & \cmark \\
Integrated Nystr\"om & \cmark & \xmark & \cmark \\
\bottomrule
\end{tabular}
\end{table}

The key differences between \texttt{TorchKM} and the two representative libraries, \texttt{scikit-learn} and \texttt{ThunderSVM}, are summarized in Table~\ref{tab:library_comparison}. Section~2 describes the core features of \texttt{TorchKM}, including GPU acceleration, support for methods beyond SVM, probability calibration, and scalable approximation for large data sets. Section~3 then outlines the exact cross-validation and spectral algorithm that make this integration possible. Performance in accuracy and time is benchmarked in Table~\ref{tab:benchmark}. Algorithmic details and further numerical results are provided in the appendix.

\section{Package Overview and Workflow}
\texttt{TorchKM} adopts a scikit-learn-style interface, so users already familiar with scikit-learn should find it easy to use. Taking the SVM as an example, the basic workflow is similar to that of \texttt{sklearn.svm.SVC}: users instantiate \texttt{TorchKMSVC}, call \texttt{fit} to train the SVM model, and then use \texttt{predict} for prediction. 

The key difference from the standard \texttt{sklearn.svm.SVC} workflow is that \texttt{TorchKMSVC} accepts a sequence of tuning parameters through \texttt{Cs}, although a single value is also permitted. In the standard scikit-learn workflow, model selection is typically handled by an outer loop over candidate values through functions such as \texttt{GridSearchCV} or \texttt{RandomizedSearchCV}, with a separate fit for each value. In \texttt{TorchKM}, by contrast, training and tuning are integrated into the algorithm itself rather than carried out through repeated external refits.

The basic usage of \texttt{TorchKMSVC} is demonstrated in the following code snippet.
\begin{pythoncode}
import numpy as np
from sklearn.datasets import make_circles
from torchkm.estimators import TorchKMSVC
X, y = make_circles(1200, factor = 0.4, noise = 0.08, random_state = 0)
# Candidate values of the tuning parameter
Cs = np.logspace(3, -3, num = 12)
# Use "cuda" to enable GPU acceleration; use "cpu" if CUDA is unavailable
clf = TorchKMSVC(kernel = "rbf", Cs = Cs, device = 'cuda', probability = True)
clf.fit(X, y) 
# Training error shown for illustration only; use held-out data in practice
clf.predict(X)
# Probability prediction via Platt scaling
clf.predict_proba(X)
# Nystrom approximation
clf.fit(X, y, low_rank = True) 
\end{pythoncode}

Setting \texttt{device="cuda"} enables GPU computation, while \texttt{device="cpu"} uses the same workflow on a CPU when a GPU is not available. For SVMs, probability estimates are available through Platt scaling via \texttt{predict\_proba}.  
By default, \texttt{low\_rank=False}, in which case \texttt{TorchKM} gives exact SVM solutions; for larger data sets, users can opt to set \texttt{low\_rank=True} to enable the Nystr\"om method and obtain approximate solutions. 

Although we use SVMs for illustration, the same interface extends to other methods, including \texttt{TorchKMDWD} for DWD \citep{marron2007distance, wang2016sparse, wang2018another}, \texttt{TorchKMLogit} for kernel logistic regression, and \texttt{TorchKMKQR} for kernel quantile regression \citep{koenker2001quantile, tang2026fastkqr}. Consequently, this shared API provides a consistent workflow across different kernel learning models.

\begin{pythoncode}
# DWD
clf_DWD = TorchKMDWD(kernel = "rbf", Cs = Cs, cv = 5, device = 'cuda')
# Logistic regression
clf_Logit = TorchKMLogit(kernel = "rbf", Cs = Cs, cv = 5, device = 'cuda')
\end{pythoncode}

\texttt{TorchKM} is released under the MIT license and distributed through GitHub and PyPI. The package is tested with \texttt{pytest}, and the repository includes detailed installation instructions, tutorials, API documentation, benchmarking notes, and a contribution guide.

\section{Core Computational Algorithms and Benchmark Performance}
In kernel learning, the full training-and-tuning pipeline is computationally expensive because both training split and tuning parameters change the linear system to be solved. In a naive $K$-fold search over $L$ candidate values, this leads to roughly $KL$ kernel solves, which require $O(KLn^3)$ operations. Due to those repeated cubic-cost operations, simply running the same algorithm on a GPU may yield only modest improvement. (See an example in Figure 1 of the appendix.) As such, the central idea behind \texttt{TorchKM} is to redesign the algorithm to shift the cubic-cost burden into matrix-vector operations. Hence the GPU’s parallel architecture is made fuller use to accelerate the whole training-and-tuning pipeline.

To achieve this, \texttt{TorchKM} employs two techniques, both of which retain exact solutions. First, an exact cross-validation formula represents each fold through a modified response vector while keeping the kernel matrix unchanged. Second, a spectral algorithm computes a single eigendecomposition of the kernel matrix and performs all subsequent updates for different tuning parameters via matrix-vector multiplications. Together, these strategies replace the aforementioned repeated cubic-cost operations with a single $O(n^3)$ decomposition followed by only $O(n^2)$ operations. Full details are provided in the appendix.

\begin{table}[t]
\centering
\caption{Comparison of \texttt{scikit-learn}, \texttt{ThunderSVM}, and \texttt{TorchKM}, averaged over 50 independent runs, with standard errors in parentheses and best values in bold. }
\label{tab:benchmark}
\renewcommand{\arraystretch}{0.8}
\setlength{\tabcolsep}{5pt}
\begin{tabular}{rrrrrrrrrrrrrrr}
\toprule
\multirow{3}{*}{$n$} & \multirow{3}{*}{$p$}
& \multicolumn{3}{c}{\texttt{scikit-learn}}
& \multicolumn{3}{c}{\texttt{ThunderSVM}}
& \multicolumn{3}{c}{\texttt{TorchKM}} \\
\cmidrule(lr){3-5} \cmidrule(lr){6-8} \cmidrule(lr){9-11}
& & \multicolumn{2}{c}{obj. values} & time (s)
  & \multicolumn{2}{c}{obj. values} & time (s)
  & \multicolumn{2}{c}{obj. values} & time (s) \\
\midrule
10000 & 10   & 0.50 & (0.01) & 1131.4 & 1.11 & (0.08) & 38.0  & \textbf{0.47} & (0.01) & \textbf{23.3} \\
      & 100  & 0.40 & (0.02) & 1974.7 & 1.04 & (0.10) & 62.7  & \textbf{0.37} & (0.02) & \textbf{17.9} \\
      & 1000 & 0.52 & (0.01) & 14322.8 & 1.00 & (0.07) & 279.4 & \textbf{0.48} & (0.01) & \textbf{37.7} \\
20000 & 10   & 0.49 & (0.01) & 3593.8 & 1.12 & (0.08) & 104.8 & \textbf{0.45} & (0.01) & \textbf{76.0} \\
      & 100  & 0.29 & (0.02) & 6206.0 & 0.49 & (0.11) & 132.9 & \textbf{0.28} & (0.01) & \textbf{78.2} \\
      & 1000 & --   & --     & $>8$h   & 0.58 & (0.07) & 580.8 & \textbf{0.47} & (0.01) & \textbf{129.3} \\
\bottomrule
\end{tabular}
\end{table}

As shown in Table~\ref{tab:benchmark}, \texttt{TorchKM} attains the lowest objective values and the shortest run times. All methods were evaluated using the same train/test splits and cross-validation folds.   All objective values were evaluated at the same tuning parameter under the same objective, equation~\eqref{eq:kernelsvm} in the appendix. Time includes the full train-and-tune pipeline. In the largest setting with \(n=20{,}000\) and \(p=1{,}000\), \texttt{scikit-learn} did not even complete within the 8-hour limit, while \texttt{TorchKM} completed the full training-and-tuning task in only 129.3 seconds. \texttt{ThunderSVM} improves over the CPU-based workflow in run time. \texttt{TorchKM} successfully further improves performance through integrated training and tuning.

\section{Conclusion}
\texttt{TorchKM} is an open-source library built on algorithm--hardware co-design, implementing exact cross-validation and spectral algorithms that naturally suit GPU computation. \texttt{TorchKM} is easy to use through a scikit-learn-style interface for SVM, DWD, logistic regression, and quantile regression. By reducing the computational overhead of tuning, \texttt{TorchKM} empowers kernel learning and supports its continued success in modern structured-data applications.

\newpage

\newpage
\section*{Appendix}
\appendix

\section{Core Algorithm}
\label{app:alg}



In this appendix, we introduce the core algorithm behind our library \texttt{TorchKM}:

\noindent
\subsection{Kernel SVM}
In this section, we use the kernel SVM as an illustration. 
Given the training data, \(\{(y_i, \mathbf{x}_i)\}_{i=1}^n\), the kernel SVM can be formulated as:
\begin{equation}
\label{eq:kernelsvm}
(\hat{\bs{\alpha}}, \hat{\beta}_{0})=\argmin_{\boldsymbol{\alpha} \in \mathbb{R}^n, \beta_0 \in \mathbb{R}}\left[\frac{1}{n} \sum_{i=1}^n \left(1 -y_i( \mathbf{K}^{\top}_{i}\bs{\alpha} + \beta_0)\right)_{+}+\lambda \boldsymbol{\alpha}^{\top} \mathbf{K} \boldsymbol{\alpha}\right],
\end{equation}
where $(1 - u)_+ = \max\{1-u, 0\}$ is the non-differentiable SVM hinge loss and \(\lambda > 0\) is a tuning parameter.

\begin{remark}
Note that \texttt{scikit-learn}'s \texttt{SVC} uses $C$ as the tuning parameter from the \texttt{libsvm} and \texttt{liblinear} convention. Thus the objective is of the form:
\[
\min_{\boldsymbol{\alpha},\, \beta_0}\;
\left[C\sum_{i=1}^n
\left(1-y_i\bigl(\mathbf K_i^{\top}\boldsymbol{\alpha}+\beta_0\bigr)\right)_+ + \dfrac{1}{2}\boldsymbol{\alpha}^\top \mathbf K \boldsymbol{\alpha} \right],
\]
where $C$ plays the inverse role of the regularization strength; $C$ and $\lambda$ are in one-to-one correspondence through \(C = \frac{1}{2n\lambda}\).
\end{remark}

To solve the optimization problem~\eqref{eq:kernelsvm}, \texttt{TorchKM} implements the finite smoothing algorithm \citep{wang2022fast} which transforms the hinge loss function into a sequence of smooth optimization problems with a \(\delta\)-smoothed hinge loss,
$$
L_\delta(u) = 
\begin{cases} 
1 - u & u \le 1-\delta, \\ \frac{1}{4 \delta}[u-(1+\delta)]^2 & 1-\delta<u<1+\delta, \\ 0 & u \ge 1+\delta,
\end{cases}
$$
and obtain the exact SVM solution. 

The smoothed problem is then solved using the proximal gradient descent. The update formula is given by:
$$
\binom{\beta_0^{(k+1)}}{\bs\alpha^{(k+1)}}
 -\binom{\beta_0^{(k)}}{\bs\alpha^{(k)}} 
= - \mathbf{H}_{\lambda}^{-1}(\mathbf{K})\binom{\mathbf{1}^\top\bs{z}^{(k)}}{\mathbf{K} \bs{z}^{(k)} + 2\lambda \mathbf{K} \bs\alpha^{(k)}},
$$
where $\mathbf{H}_\lambda(\mathbf{K}) = 2\lambda \mathbf{K} + \frac{1}{n\kappa} \mathbf{K}\mathbf{K}$ and $\bs{z}^{(k)}=(z_1, z_2, \cdots,z_n)^\top$ with each $z_i=y_iL_{\delta}^{'}[y_i (\beta_0^{(k)} + \mathbf{K}_i \bs{\alpha}^{(k)})]/n$. 

From the above update formula, we see that the main computational bottleneck in training kernel SVMs is computing $\mathbf{H}_{\lambda}^{-1}(\mathbf{K})$, which has $\mathcal{O}(n^3)$ complexity and remains costly with GPU acceleration. Additionally, cross-validation for model selection and computing solutions along the $\lambda$ path requires repeated model fitting. 

\begin{figure}[t]
    \centering
    \begin{minipage}{0.46\textwidth}
        \centering
        \includegraphics[width=\linewidth]{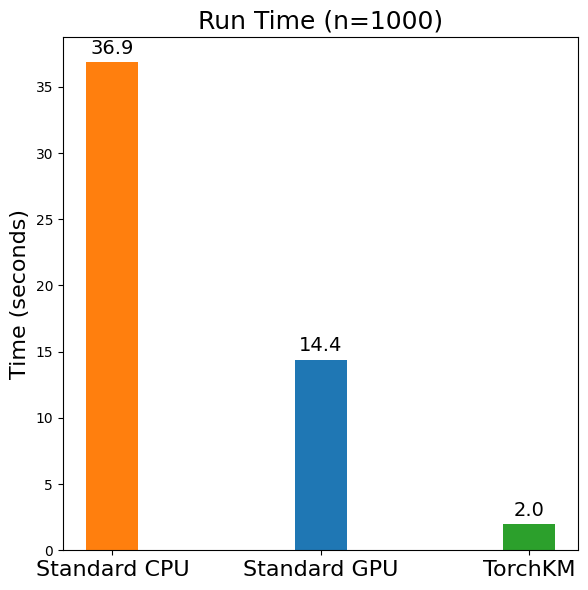}
        \caption*{(a)}
    \end{minipage}
    \hfill
    \begin{minipage}{0.46\textwidth}
        \centering
        \includegraphics[width=\linewidth]{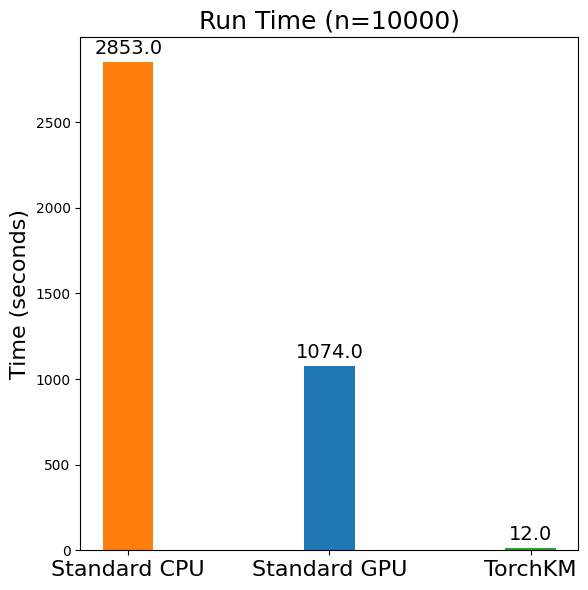}
        \caption*{(b)}
    \end{minipage}
    \caption{Run time for standard SVM training and tuning under CPU and GPU computation on simulated data sets. The standard SVM was fitted by proximal gradient descent, and model selection was performed by 5-fold cross-validation over 50 regularization parameter values. Panel (a) shows results for a data set of size $n=1{,}000$, and panel (b) shows results for a data set of size $n=10{,}000$. Lower run time indicates better computational performance. Because model fitting is repeated across cross-validation folds and along the regularization path, GPU offloading alone provides only limited speedup.}
    \label{fig:std_comparison}
\end{figure}

To illustrate the benefit of algorithm-hardware co-design, we conducted a simulation study comparing the full SVM training-and-tuning pipeline under three implementations: a standard proximal-gradient solver on CPU, the same solver on GPU, and \texttt{TorchKM}. We considered simulated data sets which are generated from a mixture-of-Gaussians model in \citet{hastie2009elements}. Let $\mu_{+}, \mu_{-} \in \mathbb{R}^{p}$ have entries equal to $\mu$ on disjoint halves of their coordinates and zero elsewhere. For each replicate, we sample centers $\mu_k^{+} \sim \mathcal{N}(\mu_{+}, I)$ and $\mu_k^{-} \sim \mathcal{N}(\mu_{-}, I)$ for $k = 1, \ldots, 10$. Positive examples are then drawn from the equal-weight mixture $\tfrac{1}{10}\sum_{k=1}^{10} \mathcal{N}(\mu_k^{+}, \sigma I)$, and negative examples from the analogous mixture on the $\mu_k^{-}$ side. We set $\mu = 2$, $\sigma = 3$, and consider $n \in \{1000, 10000\}$ with $p=10$. For each data set, we computed the solution path over 50 regularization parameters and performed model selection using 5-fold cross-validation. As shown in Figure~\ref{fig:std_comparison}, GPU acceleration of the standard solver reduces run time, but the repeated computations across cross-validation folds and tuning parameters still remain a major bottleneck. In contrast, \texttt{TorchKM} achieves substantially larger speedups by reusing matrix computations across the training-and-tuning pipeline. When \(n=10{,}000\), \texttt{TorchKM} is more than two orders of magnitude faster than the standard CPU implementation and nearly two orders of magnitude faster than the standard GPU implementation. These results show that simply offloading computation to the GPU is not sufficient; substantial acceleration requires algorithms designed to exploit GPU-friendly linear algebra. Hardware and software details are provided in Section~B.1.

To address these challenges, we introduce two strategies in the following sections: an efficient exact cross-validation method and a spectral algorithm that avoids repeated fitting when computing the solution path. Although the discussion above focuses on kernel SVM, we use it as a representative example. For kernel methods with smooth losses, since the smoothing step is unnecessary, the corresponding derivations are more direct.

\subsection{Exact Cross-Validation Formula}
Traditional cross-validation methods require the model to be retrained separately for each fold. This repeated fitting is computationally expensive and can even become infeasible when the data set is large or when leave-one-out cross-validation is required. In order to make cross-validation practical for large-scale kernel classification, \cite{wang2022fast} introduced a strategy that avoids repeatedly retraining models. Given a loss function $L(\cdot)$. Let $[v]$ denote the left-out fold with $m$ observations and define $\mathbf{K}_{[v]} \in \mathbb{R}^{(n-m)\times(n-m)}$ as the corresponding kernel matrix with the \(i\)-th rows and columns removed for \(i \in [v]\). For each tuning parameter $\lambda$, the cross-validation solution obtained by removing the set $[v]$ from the training set is given by:
$$
(\hat{\bs{\alpha}}^{[-v]}, \hat{\beta}^{[-v]}_{0})=\argmin_{\boldsymbol{\alpha} \in \mathbb{R}^{n-m}, \beta_0 \in \mathbb{R}}\left[\frac{1}{n-m} \sum_{j \notin[v]} L\left(y_j( \mathbf{K}^{\top}_{[v],j}\bs{\alpha} + \beta_0)\right)+\lambda \boldsymbol{\alpha}^{\top} \mathbf{K}_{[v]} \boldsymbol{\alpha}\right].
$$

Instead of refitting a separate model for each fold, the optimization is reformulated so that the same kernel matrix can be reused, with only minor modifications to the response vector. In particular, define $\tilde{\bs{y}}^{[v]}$ by letting $\tilde{y}_j^{[v]}=0$ for $j \in v$ and $\tilde{y}_j^{[v]}=y_j$ for all $j \notin[v]$, and then by Lemma 2 of \cite{wang2022fast}, it holds that
$$
(\tilde{\bs{\alpha}}^{[-v]}, \hat{\beta}^{[-v]}_{0})=\argmin_{\boldsymbol{\alpha} \in \mathbb{R}^n, \beta_0 \in \mathbb{R}}\left[\frac{1}{n-m} \sum_{j=1}^n L\left(\tilde{y}_j^{[v]}( \mathbf{K}^{\top}_{j}\bs{\alpha} + \beta_0)\right)+\lambda \boldsymbol{\alpha}^{\top} \mathbf{K} \boldsymbol{\alpha}\right].
$$
and $\hat{\bs{\alpha}}^{[-v]}$ can be obtained from $\tilde{\bs{\alpha}}^{[-v]}$ by removing the entries corresponding to the held-out observations. 
Consequently, the exact cross-validation formula allows us to work with the same kernel matrix and slightly modified \(\tilde{y}_j^{[v]}\). We compute and store the inversion of \(\mathbf{H}_\lambda(\mathbf{K})\) only once, avoiding the need to invert the matrix \( V \) times.

\begin{algorithm}[t]
\caption{Exact leave-one-out cross-validation for kernel SVM in \texttt{TorchKM} (\texttt{cvksvm})}
\begin{algorithmic}[1]
\REQUIRE{$\bs{y}, \mathbf{K}$, and $\lambda$ sequence}
\STATE{Initialize $\delta$. The smooth function is $L_{\delta}$. Let $\kappa = 2\delta$.}
\STATE{Initialize $\tilde{\bs{\alpha}}^{[-i]} = \mathbf{0}$ for $i = 1,\dots,n$}.
\STATE Compute eigendecomposition $\mathbf{K} = \mathbf{U} \mathbf{\Lambda} \mathbf{U}^\top$
\FOR{each tuning parameter $\lambda_{\ell}$, $\ell = 1, 2, \ldots, L$,}
    \REPEAT
    \STATE Compute $\bs\Pi_{\lambda_{\ell}} = 2\lambda_{\ell}\bs\Lambda + 1/(n\kappa)\bs\Lambda\bs\Lambda$.
    \STATE Compute $\mathbf{v}=\mathbf{U}\bs{\Lambda}\bs\Pi_{\lambda_{\ell}}^{-1}\mathbf{U}^{\top}\mathbf{1}$
and $g=1/\bigl(n\mathbf{1}^{\top}\mathbf{U}\bs{\Lambda}\bs\Pi_{\lambda_{\ell}}^{-1} \bs{\Lambda}\mathbf{U}^{\top}\mathbf{1}\bigr)$. 
\FOR {$i=1, \dots, n$}
    \STATE Let $\tilde{y}_j^{[-i]}=y_j$ if $j \neq i$, and $\tilde{y}_i^{[-i]}=0$.
    \STATE Set 
    $(\bar{\bs{\alpha}},\bar{\beta}_0)
    \leftarrow
    (\tilde{\bs{\alpha}}^{[-i]},\tilde{\beta}_0^{[-i]})$
    and 
    $(\bs{\alpha}^{\prime},\beta_0^{\prime})
    \leftarrow
    (\tilde{\bs{\alpha}}^{[-i]},\tilde{\beta}_0^{[-i]})$.
    \STATE Let $r=1$.

    \REPEAT 
        \STATE Compute $r^{\prime}=\frac{1+\sqrt{1+4 r^2}}{2}$.
        \STATE Update
        $$
        (\bar{\bs{\alpha}},\bar{\beta}_0)
        \leftarrow
        (\tilde{\bs{\alpha}}^{[-i]},\tilde{\beta}_0^{[-i]})
        +
        \frac{r-1}{r^{\prime}}
        \left\{
        (\tilde{\bs{\alpha}}^{[-i]},\tilde{\beta}_0^{[-i]})
        -
        (\bs{\alpha}^{\prime},\beta_0^{\prime})
        \right\}.
        $$
        \STATE Let $\bar{\bs{z}}=\left(\bar{z}_1, \ldots, \bar{z}_n\right)^{\top}$, with
        $ \bar{z}_j = \tilde{y}_j^{[-i]} L_\delta^{\prime}
        [\tilde{y}_j^{[-i]}
        (\mathbf{K}_j^{\top}\bar{\bs{\alpha}} + \bar{\beta}_0)]/n.$
        \STATE From right to left avoiding matrix multiplications, compute
        $$
        \Delta\bs\alpha = g(-(\mathbf{1}^\top \bar{\mathbf{z}})\mathbf{v} + \mathbf{vv}^\top \mathbf{K}(\bar{\mathbf{z}} + 2\lambda_{\ell}\bar{\bs\alpha}) ) + \mathbf{U}\bs{\Lambda}\bs\Pi_{\lambda_{\ell}}^{-1}\mathbf{U}^{\top}(\bar{\mathbf{z}} + 2\lambda_{\ell}\bar{\bs\alpha}).
        $$
        \STATE Compute
        $\Delta \beta_0 = g(\mathbf{1}^\top \bar{\mathbf{z}}) - g \mathbf{v}^\top \mathbf{K}(\bar{\mathbf{z}} + 2\lambda_{\ell}\bar{\bs\alpha})$
        \STATE Update 
        $(\bs{\alpha}^{\prime},\beta_0^{\prime})
        \leftarrow
        (\tilde{\bs{\alpha}}^{[-i]},\tilde{\beta}_0^{[-i]})$.
        \STATE Update 
        $
        (\tilde{\bs{\alpha}}^{[-i]},\tilde{\beta}_0^{[-i]})
        \leftarrow
        (\bar{\bs{\alpha}}-\Delta\bs{\alpha},
        \bar{\beta}_0-\Delta\beta_0).
        $
        \STATE Update $r \leftarrow r^{\prime}$.
    \UNTIL the convergence condition is met.
    \ENDFOR
    \STATE Shrink $\delta = \eta\delta$, where $\eta = 0.125$.
 Update $L_{\delta}$ and $\kappa = 2\delta$.
    \UNTIL the KKT conditions of all SVM models are satisfied.
\ENDFOR
\end{algorithmic}
\end{algorithm}

\subsection{Spectral Algorithm}
Even with the exact cross-validation formula, we still need to solve the optimization problem along the solution path for each tuning parameter $\lambda$. Recall that the kernel SVM training updates take the following form:
$$
\binom{\beta_0^{(k+1)}}{\bs\alpha^{(k+1)}}
 -\binom{\beta_0^{(k)}}{\bs\alpha^{(k)}} 
= - \mathbf{H}_{\lambda}^{-1}(\mathbf{K}) \mathbf{d}^{(k)}, \ \text{ where } \mathbf{d}^{(k)} = \binom{\mathbf{1}^\top\bs{z}^{(k)}}{\mathbf{K} \bs{z}^{(k)} + 2\lambda \mathbf{K} \bs\alpha^{(k)}}
$$
and $\mathbf{H}_\lambda(\mathbf{K}) = 2\lambda \mathbf{K} + \frac{1}{n\kappa} \mathbf{K}\mathbf{K}$.
Each step would require a new inversion of $\mathbf{H}_{\lambda}(\mathbf{K})$ for every $\lambda$, whose complexity is $O(n^3)$.

To address this, \texttt{TorchKM} applies a spectral algorithm: compute the eigendecomposition \(\mathbf K=\mathbf U\mathbf\Lambda\mathbf U^\top\) first. For each $\lambda$, define
\(\mathbf\Pi_\lambda=2\lambda \mathbf\Lambda+\frac{1}{n\kappa}\mathbf{\Lambda\Lambda}.\) Then $\mathbf H_\lambda^{-1}$ can be written as:
\[
    \mathbf H_\lambda^{-1}(\mathbf K)
    =
    g
    \begin{pmatrix}
    1\\
    -\mathbf v
    \end{pmatrix}
    \begin{pmatrix}
    1 & -\mathbf v^\top
    \end{pmatrix}
    +
    \begin{pmatrix}
    0 & \mathbf 0^\top\\
    \mathbf 0 & \mathbf U\mathbf\Pi_\lambda^{-1}\mathbf U^\top
    \end{pmatrix},
\]
where
$\mathbf{v}=\mathbf{U}\bs{\Lambda}\Pi_{\lambda}^{-1}\mathbf{U}^{\top}\mathbf{1}$, and $g=1/\bigl(n\mathbf{1}^{\top}\mathbf{U}\bs{\Lambda}\Pi_{\lambda}^{-1} \bs{\Lambda}\mathbf{U}^{\top}\mathbf{1}\bigr)$. At the $k$th iteration, we compute \(\mathbf H_\lambda^{-1}(\mathbf K)\mathbf d^{(k)}\) from right to left with only $O(n^2)$ complexity.

Consequently, after a single \(O(n^3)\) eigendecomposition of \(\mathbf K\), the subsequent computations along the \(\lambda\)-path reduce to \(O(n^2)\) matrix--vector multiplications. This structure is well suited to GPU parallelism and is a key reason for the computational efficiency of \texttt{TorchKM}.

\subsection{The \texttt{TorchKM} Algorithm}
With the exact cross-validation formula and the spectral algorithm, we implement the \texttt{TorchKM} algorithm as detailed in Algorithm 1. For illustration, we present the leave-one-out cross-validation version for kernel SVM over a sequence of tuning parameters \(\lambda_1, \lambda_2, \ldots, \lambda_L\). To further accelerate computation along the regularization path, we use warm starts: the solution obtained at \(\lambda_{\ell}\) is used as the initial value for solving the problem at \(\lambda_{\ell+1}\). We also incorporate Nesterov's acceleration \citep{nesterov1983method} to speed up convergence.

Algorithm 1 presents the leave-one-out case for notational simplicity. For $V$-fold cross-validation, the held-out index $i$ is replaced by a held-out fold and the same spectral quantities can be used to compute the fold-wise validation loss values.

\subsection{Nystr\"om Kernel Approximation}

So far, we have introduced \texttt{TorchKM} which provides exact solution in the full-kernel setting. For large-scale problems, \texttt{TorchKM} also supports the Nystr\"om approximation. Working directly with a full \(n \times n\) kernel matrix quickly becomes infeasible when \( n \) is large. The Nystr\"om method addresses this issue by constructing a compressed representation of the kernel using only a subset of the training points. 
\texttt{TorchKM} adopts a customized Nystr\"om kernel approximation framework that avoids recomputing the approximation for every regularization parameter and every cross-validation fold, as is typically required in conventional Nystr\"om methods. \texttt{TorchKM} can perform a single unified Nystr\"om approximation of $\mathbf{K}$ outside the $\lambda$-path and cross-validation loops, making model training efficient.

\subsection{Multiclass Classification}
Unlike \texttt{ThunderSVM} which handles multiclass classification \citep{wen2018efficient}, the current version of \texttt{TorchKM} only focus on binary classification, and multiclass problems needs to be handled externally through the standard one-vs-rest or one-vs-one approaches. Native multiclass support is a natural direction for future development; for example, multicategory kernel DWD provides a potential route
for those kernel machines \citep{wang2019multicategory}.

\begin{figure}[t]
    \centering
    \includegraphics[width=0.6\linewidth]{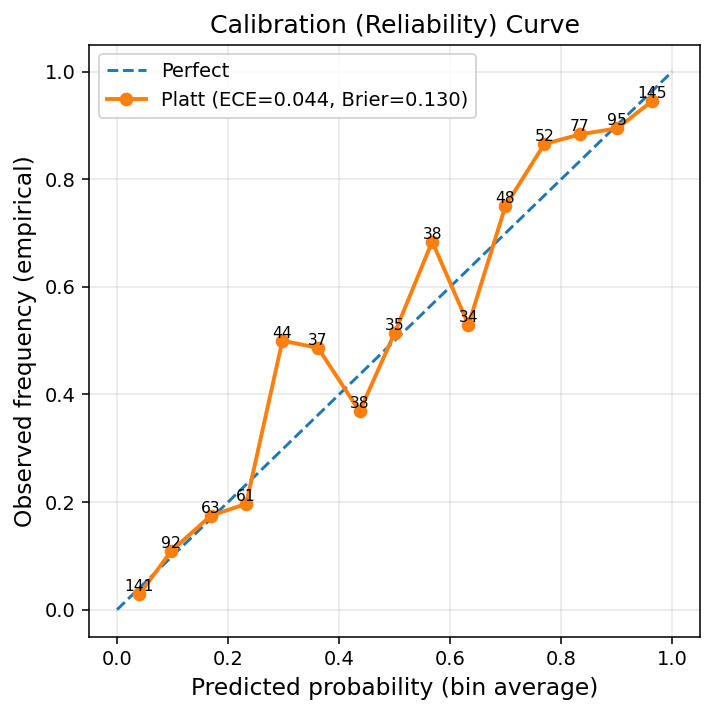}
    \caption{Reliability curve for Platt-calibrated probabilities on the independent test set. The plotted curve compares observed frequencies with predicted probabilities, and the dashed 45$^\circ$ line denotes perfect calibration. Curves closer to the diagonal indicate better calibration. The expected calibration error (ECE) is 0.044 and the Brier score is 0.130; smaller values indicate better calibration, suggesting good overall agreement between predicted probabilities and observed frequencies.}
    \label{fig:platt}
\end{figure}

\noindent
\subsection{Platt Scaling Plot}
\texttt{TorchKM} also provides calibrated probability estimates in addition to class predictions. Platt scaling is a probability calibration method that transforms a model’s raw decision scores into estimated class probabilities by fitting a sigmoid function. Its main purpose is to improve the interpretability and reliability of the classifier. In this section, we used the same binary classification setting as in the previous section. The training set consisted of $n=10000 $ observations, and the test set consisted of 1000 observations, each with $p=10$ predictors. The RBF-kernel SVM was fitted over a sequence of 50 regularization parameter values. The Platt calibration plot was then used to evaluate how well the calibrated probabilities align with the observed outcome frequencies. We evaluated the calibration performance on an independent test set using a reliability curve with the expected calibration error (ECE), and the Brier score. Figure~\ref{fig:platt} shows the reliability curve for Platt-calibrated probabilities on the independent test set. The dashed line denotes perfect calibration. The curve remains close to the identity line overall, with ECE of 0.044 and Brier score of 0.130, which indicates good overall calibration.

\section{Additional Experiment Results}

\subsection{Experimental Setup}
\paragraph{Hardware and software environment.}
All timing experiments reported in Tables~\ref{tab:benchmark}--\ref{tab:dnn-cvknyssvm-nystrom} were carried out on a single workstation equipped with one NVIDIA L40S GPU with 48 GB of memory, an AMD EPYC 9334 (32-Core) CPU, and 768 GB of system RAM,
running Ubuntu 22.04 with CUDA 12.1. \texttt{TorchKM} was executed under Python~3.11 with PyTorch 2.4.1 (CUDA build). The \texttt{scikit-learn} baseline used
\texttt{scikit-learn} 1.1.3 together with
\texttt{NumPy} 1.25.2 and \texttt{SciPy} 1.9.3 on the CPU.
\texttt{ThunderSVM} was built from the official release 0.3.4
against CUDA 12.1 and run on the same GPU as \texttt{TorchKM}.

All run times are wall-clock times measured end-to-end for the full training-and-tuning pipeline, averaged over 50 runs in Table~\ref{tab:benchmark} and 10 runs in Tables~\ref{tab:dnn-cvksvm-thundersvm}--\ref{tab:dnn-cvknyssvm-nystrom}.

 \begin{table}[t]
\centering
\caption{Classification accuracy (Acc) and run time (Time, in seconds) for  \texttt{TorchKM} and \texttt{ThunderSVM} on the \texttt{a7a}, \texttt{a8a}, and \texttt{w7a} benchmark classification data sets. Numbers in parentheses give the sample size of each data set. Both \texttt{TorchKM} and \texttt{ThunderSVM} use the full-kernel SVM solver without a Nystr\"om approximation. Higher accuracy and lower run time indicate better performance. Across the three data sets, \texttt{TorchKM} attains better or matched accuracy and shorter run time.}
\begin{tabular}{ccccccc}
\hline
\multirow{2}{*}{\textbf{Data (size)}} && \multicolumn{2}{c}{\textbf{TorchKM}} && \multicolumn{2}{c}{\textbf{ThunderSVM}}\\
\cline{3-4} \cline{6-7}
 & & Acc & Time (s) && Acc & Time (s)\\
\hline
a7a (16{,}100) && 0.830 & 48.53 && 0.828 & 108.50\\
a8a (22{,}696) && 0.833 & 106.92 && 0.829 & 154.28\\
w7a (24{,}692) && 0.971 & 135.06 && 0.971 & 203.11\\
\hline
\end{tabular}
\label{tab:dnn-cvksvm-thundersvm}
\end{table}

\paragraph{Tuning-parameter grid and cross-validation.}
For all libraries, we used a grid of \(50\) candidate regularization values,
spaced log-uniformly over \([-3,3]\), corresponding to
\[
C \in [10^{-3}, 10^{3}]
\]
under the \texttt{scikit-learn}/LIBSVM parameterization. Equivalently, one may
use the \(\lambda\)-parameterization by converting each value of \(C\) via
\[
C = \frac{1}{2n\lambda}.
\]
For \texttt{TorchKM}, model selection in
Tables~\ref{tab:benchmark}--\ref{tab:dnn-cvknyssvm-nystrom} was performed using
\(10\)-fold cross-validation. The competing libraries were also tuned using
\(10\)-fold cross-validation via \texttt{GridSearchCV} or \texttt{cross\_val\_score}.
 
\paragraph{Benchmark data.}
Tables~\ref{tab:dnn-cvksvm-thundersvm} and~\ref{tab:dnn-cvknyssvm-nystrom} use standard benchmark classification data sets from the LIBSVM repository, available at
\url{https://www.csie.ntu.edu.tw/~cjlin/libsvmtools/datasets/}.
The specific files we used are:
\begin{itemize}
  \item \textbf{a7a, a8a, a9a} (UCI Adult):\\
        \url{https://www.csie.ntu.edu.tw/~cjlin/libsvmtools/datasets/binary/a7a},\\
        \url{https://www.csie.ntu.edu.tw/~cjlin/libsvmtools/datasets/binary/a8a},\\
        \url{https://www.csie.ntu.edu.tw/~cjlin/libsvmtools/datasets/binary/a9a}.
  \item \textbf{w7a, w8a} (Web Page):\\
        \url{https://www.csie.ntu.edu.tw/~cjlin/libsvmtools/datasets/binary/w7a},\\
        \url{https://www.csie.ntu.edu.tw/~cjlin/libsvmtools/datasets/binary/w8a}.
  \item \textbf{ijcnn1}:\\
        \url{https://www.csie.ntu.edu.tw/~cjlin/libsvmtools/datasets/binary/ijcnn1.bz2}.
  \item \textbf{covtype.binary}:\\
        \url{https://www.csie.ntu.edu.tw/~cjlin/libsvmtools/datasets/binary/covtype.libsvm.binary.scale.bz2}.
  \item \textbf{MNIST8m (4 vs.~6)}: extracted as the subset of digits~4 and~6 from
        \texttt{mnist8m.scale.xz} at\\
       \url{https://www.csie.ntu.edu.tw/~cjlin/libsvmtools/datasets/multiclass/mnist8m.scale.xz};
        see also \url{https://leon.bottou.org/projects/infimnist}.
\end{itemize}

Each data set was used in its standard LIBSVM-format release without further preprocessing beyond feature scaling to $[-1,1]$ where provided.

\subsection{Benchmark Results}
In this section, we demonstrate the performance of \texttt{TorchKM} in terms of classification accuracy and run time on three  benchmark data sets: \texttt{a7a}, \texttt{a8a}, and \texttt{w7a}, which contain 16,100, 22,696, and 24,692 samples, respectively. Table~\ref{tab:dnn-cvksvm-thundersvm} shows that \texttt{TorchKM} consistently improves predictive accuracy while significantly reducing run time. This demonstrates the advantage of \texttt{TorchKM} as an efficient and effective kernel learning approach.

For the larger-scale evaluation, we used five benchmark data sets: \texttt{a9a}, \texttt{w8a}, \texttt{ijcnn}, \texttt{covtype}, and \texttt{MNIST8m} (4 vs 6), which contain 32,561, 49,749, 49,990, 581,012, and 1,270,178 samples, respectively. These data sets were selected to extend the comparison to larger classification problems and to examine the scalability of the solvers. We evaluated both the predictive performance and the tuning efficiency of \texttt{TorchKM} with Nystr\"om approximation, and Nystr\"om implemented in \texttt{scikit-learn}.

Table \ref{tab:dnn-cvknyssvm-nystrom} presents the comparison of these solvers on the larger benchmark data sets in terms of accuracy and run time. The improvement of \texttt{TorchKM} becomes even more pronounced on \texttt{MNIST8m} (4 vs 6), where it obtains the higher accuracy of 0.997 in just 64.647 seconds, compared with 0.996 in 5189.03 seconds for \texttt{scikit-learn} with Nystr\"om. These results demonstrate that \texttt{TorchKM} not only delivers consistently superior accuracy, but also scales far more efficiently on large-scale data sets.

\begin{table}[t]
\centering
\caption{Classification accuracy (Acc) and run time (Time, in seconds) for \texttt{cvknyssvm} in \texttt{TorchKM} and the Nystr\"om approximation in \texttt{scikit-learn} on the \texttt{a9a}, \texttt{w8a}, \texttt{ijcnn}, \texttt{covtype}, and \texttt{MNIST8m} (4 vs 6) benchmark classification data sets. Numbers in parentheses give the sample size of each data set. Higher accuracy and lower run time indicate better performance. Across all five data sets, \texttt{TorchKM} attains higher accuracy and shorter run time, with the largest gain appearing on \texttt{MNIST8m} (4 vs 6).}
\begin{tabular}{ccccccc}
\hline
\multirow{2}{*}{\textbf{Data (size)}} && \multicolumn{2}{c}{\textbf{TorchKM}} & &\multicolumn{2}{c}{\textbf{Nystr\"om (sklearn)}}\\
\cline{3-4} \cline{6-7}
 & &Acc & Time (s) && Acc & Time (s)\\
\hline
a9a (32{,}561)      &&0.851 &17.86  
&&0.849 &205.83  \\
w8a (49{,}749)      & &0.981 &20.78  &&0.976 &275.37  \\
ijcnn1 (49{,}990)    &&0.982 &27.42  &&0.977  &275.49  \\
covtype (581{,}012) & &0.807  &31.11  &&0.786  &2269.35  \\
MNIST8m (1{,}270{,}178)   & &0.997 &64.65  &&0.996  &5189.03  \\
\hline
\end{tabular}
\label{tab:dnn-cvknyssvm-nystrom}
\end{table}

\vskip 0.2in
\bibliography{torchKM}

\end{document}